\documentclass[
reprint,
amsmath,amssymb,
aps,
apl,
superscriptaddress 
]{revtex4-2}

\usepackage{graphicx}
\usepackage{dcolumn}
\usepackage{bm}
\usepackage{xcolor}
\usepackage{hyperref}
\hypersetup{colorlinks=true,allcolors=blue!70!black}
\usepackage{float}

\begin{document}
	
	\preprint{APS/123-QED}
	
	\title{Photonic reservoir computing with dimensionally compressed readout}

	\author{Gérald Kobi}
	\email{gerald.kobi@centralesupelec.fr}
	\affiliation{Université de Lorraine, CentraleSupélec, LMOPS, F-57000 Metz, France}
	\affiliation{Instituto de F\'isica Interdisciplinar y Sistemas Complejos (IFISC), CSIC--UIB, Campus UIB, E-07122 Palma de Mallorca, Spain}
	
	\author{Mohab Abdalla}
	\affiliation{Université de Lorraine, CentraleSupélec, LMOPS, F-57000 Metz, France}	
	
	\author{Miguel C. Soriano}
    \affiliation{Instituto de F\'isica Interdisciplinar y Sistemas Complejos (IFISC), CSIC--UIB, Campus UIB, E-07122 Palma de Mallorca, Spain}

	\author{Damien Rontani}
    \email{damien.rontani@centralesupelec.fr}
	\affiliation{Université de Lorraine, CentraleSupélec, LMOPS, F-57000 Metz, France}
	
	\date{\today}
	
	\begin{abstract}
		
		This work addresses a hardware constraint in reservoir computing: the limited size of the readout layer imposed by systems with a physical readout. We investigate a strategy to accommodate this constraint based on random projection, which compresses high-dimensional reservoir states into a lower-dimensional subspace while preserving key properties of the source space and information-processing capabilities. To evaluate this approach, we compare a small, standalone time delay reservoir against a larger configuration whose output is projected down to match the same restricted readout dimension. Using task-independent metrics, we demonstrate that the distribution of information-processing capacities may differ between the two configurations, even at identical readout sizes. Furthermore, we perform a comprehensive hyperparameter scan to assess how both systems behave under varying physical regimes. Finally, we benchmark this approach on the standard NARMA10 task, showing that the random projection framework can yield superior performance compared to a standalone constrained reservoir, within specific compression range. These results provide a scalable pathway to bypass physical readout bottlenecks in hardware-based reservoir computing.

	\end{abstract}

	\maketitle

	\section{\label{sec:level1}Introduction}
	
	Reservoir Computing (RC) has emerged as a powerful framework for solving complex temporal tasks \cite{verstraeten2007experimental}, using remarkably simple learning rules that avoid the computational burden of traditional backpropagation-based training \cite{difficult_retropropragation_gradinet}. As originally proposed by Jaeger \cite{jaeger2001echo}, the core of RC lies in utilizing a fixed, randomly initialized \textit{reservoir} \cite{Schrauwen_2007, Tino_rodan}, a high-dimensional dynamical system that maps input signals into a complex state space, while training only the linear output layer\cite{LUKOSEVICIUS2009127}. 
	
	Over the last two decades, this paradigm has undergone a pivotal shift, moving from abstract digital simulations to a diverse array of physical hardware \cite{Vandoorne:08,TANAKA2019100}. This transition represents a major leap, transitioning from digital implementation to high-performance analog computing. While these physical reservoir computers are not intended to replace traditional general-purpose architectures, they offer several advantages for specific tasks, such as real-time signal processing \cite{Vandoorne:08} or complex pattern recognition \cite{Rafayelyan_2020}, by providing ultra-high-speed performance \cite{High_Speed_Photonic_Reservoir} and superior power efficiency \cite{Feldmann_2021}. Among these modalities, photonic implementations have recently taken center stage \cite{Abdalla_2026}, making use of the high bandwidth and massive parallel processing capabilities inherent to optical systems \cite{phtoonic_secodn_review}.
	
	To overcome scalability challenges associated with large-scale spatial networks \cite{Photonic_vandoorne}, the field has extensively explored the Time-Delay Reservoir Computing (TDRC) architecture, originally introduced by Appeltant et al.\cite{Appeltant2011}. By employing a single non-linear node sampled at fixed time intervals, this approach maps temporal information into a high-dimensional virtual state space. This multiplexing strategy enables the emulation of complex neural dynamics with a minimal hardware footprint, while remaining compatible with high-speed photonic components \cite{Brunner2013,Vatin:18}.

	In conventional Physical Reservoir Computing (PRC), readout layers typically rely on digital processing to ensure full accessibility to internal states. However, it introduces a major latency bottleneck in photonic systems, where optoelectronic conversion exceeds the reservoir's intrinsic timescale. Even advanced schemes featuring real-time online training \cite{Antonik_2017} remain constrained by these electronic loops. To bypass this, analog readout layers perform matrix computations directly within the optical domain \cite{optical_output_first}. By leveraging hardware for analog optical weighting \cite{Tait_2017,Meng2021}, these frameworks eliminate continuous analog-to-digital conversion, minimizing both latency and energy consumption.
	
	However, the transition toward fully analog systems, particularly in integrated photonics, has introduced significant detection constraints that limit the ability to exploit the reservoir's high-dimensional state space. 	 
	
	Therefore, these architectures remain experimentally challenging to fabricate and operate, with current state-of-the-art implementations restricted to few tens of readout states \cite{state_of_art_outtput_physical_layer, VanAssche2026}. Such a limitation is significant, as theoretical analyses of "vanilla" RC suggest that restricting the number of neurons of a system might affect the overall performance negatively \cite{LUKOSEVICIUS2009127}. 
	
	Consequently, a compelling motivation is to leverage system performance under the constraints of a limited output layer. Indeed, it remains largely unexplored how a compressed-readout reservoir (projecting a large-dimensional state of size $N$ onto a small-dimensional output of size $M$) compares to a fully accessible, natively small-dimensional reservoir (of size $M$). Addressing this gap, this work focuses on Time-Delay Reservoir Computers (TDRCs) to investigate how random projection, as a dimensionality reduction scheme, impacts system performance.

	We first analyze the processing capacity of these systems within the context of dimensionality reduction. By evaluating our findings through the lens of task-independent metrics, we examine how this capacity evolves across different configurations as a function of the physical hyperparameters. Finally, we consolidate these insights to demonstrate how tuning these parameters can yield a significant performance enhancement on a nonlinear regression task.
	
	\section{\label{sec:level2}Scope of the Study: Methodology and Metrics}
	
	This section outlines our methodology, starting with the photonic systems under consideration: reservoir computing architectures based on the optoelectronic Ikeda model. We then describe the particular dimension reduction technique implemented and provide a detailed overview of the specific task-independent metrics employed.

	\subsection{Ikeda system}

	In this study, we use an Ikeda-based reservoir characterized by a $\sin^2$ nonlinearity. The system dynamics under the influence of an external input signal are governed by the following delay differential equation:
	\begin{equation}
		\dot x(t) T_{r}=-x(t) + \beta \sin^2(x(t-\tau)+\gamma m(t) + \phi_{0}).
	\end{equation}
	Here, $x$ represents the state variable, $\beta$ denotes the feedback strength, and $T_{r}$ is the response time. The parameter $\tau$ represents the characteristic delay time of the system, which is physically determined by components such as the length of a fiber-optic loop \cite{Larger:12}. The input term is defined as $m(t) = u(t) \cdot \text{mask}(t)$, where $u(t)$ is the raw input signal. The $\text{mask}(t)$ is a piecewise constant function that maps the input onto $N$ virtual nodes, updating every $\theta = T_{in}/N$. In this context, $T_{in}$ denotes the input injection period, representing the duration for which each entry $u(t)$ is held. Unless stated otherwise, the following default parameter values are adopted throughout this study: $T_{r} = 0.4\theta$, $\beta = 1.6$, $\gamma = 0.01$, $\phi_{0} = 0.2$, and $\tau = T_{\text{in}} \times 0.9$.
	
\begin{figure}[H]
		\centering
		\includegraphics[width=\columnwidth]{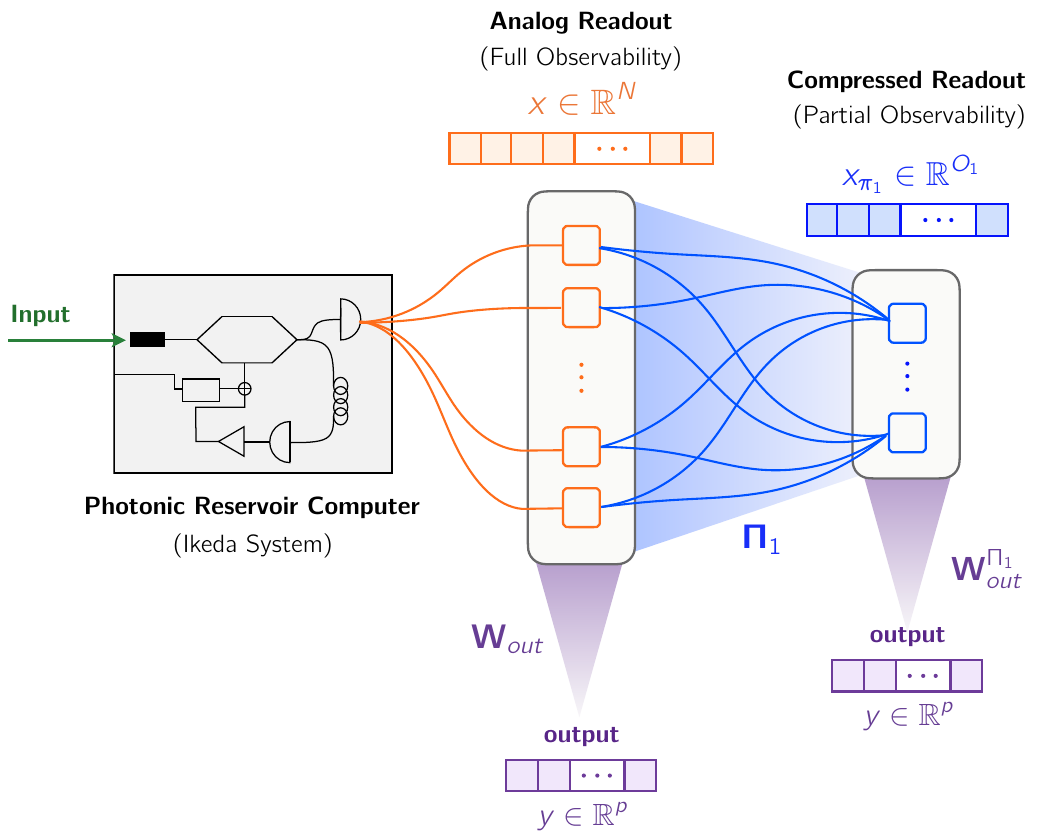}
		\caption{\textbf{Projection scheme on a Time Delay Reservoir.} On the left, an input signal is injected into a photonic reservoir computer (Ikeda system) to generate complex dynamical states. In the middle, the full set of these states is collected as a high-dimensional analog readout vector $x \in \mathbb{R}^N$ (Full Observability) to map the output $y \in \mathbb{R}^p$ via $W_{out}$. In blue, a dimensionality reduction projection $\Pi_1$ is applied to compress the system. On the right, the resulting lower-dimensional compressed readout $x_{\pi_1} \in \mathbb{R}^{O_1}$ (Partial Observability) is collected to compute the optimized target output via $W_{out}^{\Pi_1}$.}
		\label{fig:wide1}
	\end{figure}

	Originally it was implemented via a Mach-Zehnder interferometer, an optical fiber as the delay line and a detector made of a single photodiode\cite{Larger:12}. In our specific configuration, we introduce a layer between the reservoir's output and the final system output. This intermediate stage is designed to reduce the dimensionality of the reservoir states prior to the readout phase. The entire system follows the scheme illustrated in Fig.~\ref{fig:wide1}.

	\subsection{Projection method presentation}
	
	For the dimensionality reduction illustrated in Fig.~\ref{fig:wide1}, we explore a specific method, known as random projection (RP). This method is based on the Johnson-Lindenstrauss lemma \cite{johnson1984extensions} and its matrix-based application proposed by Dimitris Achlioptas \cite{ACHLIOPTAS2003671}. The lemma states the following:\newline Let $A \in \mathbb{R}^{n \times d}$ be a matrix containing $n$ points as its rows. For $\varepsilon, \lambda > 0$, let $k \ge \frac{4+2\lambda}{\varepsilon^2/2 - \varepsilon^3/3} \log(n)$. Define $E = \frac{1}{\sqrt{k}} AR$, where $R \in \mathbb{R}^{d \times k}$ contains entries $r_{ij}$ drawn i.i.d. from:$$r_{ij} = \sqrt{3} \times \begin{cases} +1 & \text{w.p. } 1/6 \\ 0 & \text{w.p. } 2/3 \\ -1 & \text{w.p. } 1/6 \end{cases}$$With probability at least $1 - n^{-\lambda}$, the linear mapping $f(x) = \frac{1}{\sqrt{k}} x R$ preserves the pairwise distances for any two rows $u,v$ of $A$:\begin{equation}(1-\varepsilon)|u-v|^2 \le |f(u)-f(v)|^2 \le (1+\varepsilon)|u-v|^2.\end{equation} 
	Essentially, this theorem guarantees that a random projection can significantly reduce the dimensionality of vectors within a given state space without distorting the underlying geometric relationships (distances) of the original data.
	Since the projection is a linear process (implemented in this framework via matrix multiplication) applying this scheme to the readout layer yields linear combinations of the internal reservoir states, which we hereafter refer to as observables ($O$). This property is particularly advantageous for constrained readouts, as the theorem ensures that a highly sparse projection matrix suffices to sample reservoir states with minimal information loss.
	
\begin{figure}[H]
		\centering
		\includegraphics[width=\columnwidth]{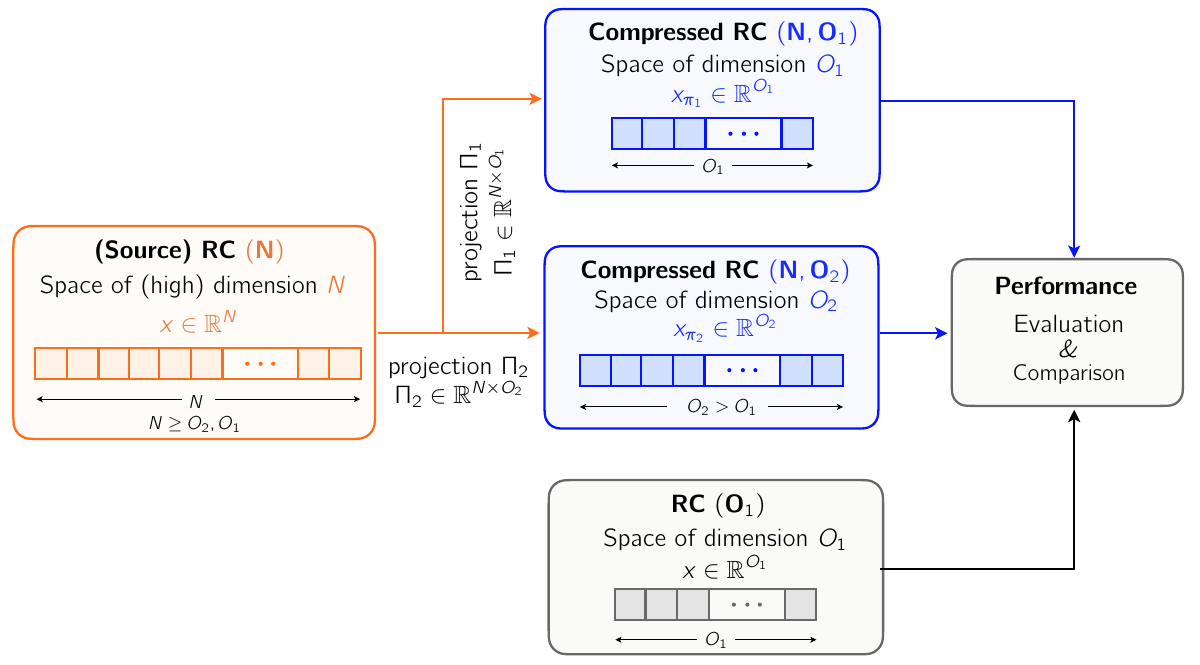}
		\caption{Projection terminology and notation: The source reservoir is denoted by $(N_o)$ (with size $N = N_o$), while the unprojected baseline case is denoted by the single-entry tuple $(O_j)$ (where $N = O = O_j$). The projected configurations, where the high-dimensional source is mapped onto a lower-dimensional readout, are represented by the tuple $(N_o, O)$. Two distinct projected cases are shown: $(N_o, O_i)$ and $(N_o, O_j)$ (with $i \neq j$), corresponding to output dimensions of $O = O_i$ and $O = O_j$, respectively.}
		\label{fig:wide2}
	\end{figure}

	We define two distinct configurations derived from a primary reservoir system of $N$ neurons. The standard, unprojected case is denoted by the single-entry tuple $(N)$, where no projection is applied and the observables are simply all the reservoir neurons ($O = N$). Conversely, the projected case is denoted by the tuple $(N, O)$, where a random projection is applied to the high-dimensional state matrix to obtain a reduced matrix of $O$ observables. Throughout this work, the dimensionality reduction constraint $O \le N$ is maintained. These definitions and the underlying notation are visually summarized in Fig.~\ref{fig:wide2}.

	\subsection{Metrics used to compare the different cases }
	
	We will analyze the influence of random projection on constrained readout layer size using task-independent metrics, which allow us to evaluate the system's intrinsic performances without biasing the results toward specific tasks. The first metric, widely adopted in the reservoir computing community, is the Memory Capacity ($MC$). Introduced by Jaeger in \cite{jaeger:techreport2002}, this metric quantifies the system's ability to reconstruct past inputs from its current state.
    For a task indexed by \(k\), $C(y_k)$ measures the system's ability to reconstruct a specific target function $y_k$ from its current state, calculated as the normalized squared correlation between $y_k$ and its estimate $\widehat{y}_k$:
	\begin{equation}
		C(y_k) = \frac{\langle y_k, \widehat{y}_k \rangle^2}{\langle y_k, y_k \rangle \langle \widehat{y}_k, \widehat{y}_k \rangle}.
	\end{equation}
    Consider an infinite number of tasks, each of which is to construct the input \(k\) steps into the past. The sum of all constituent capacities yields the \(MC\).
    For an input $u$ drawn from a uniform distribution $u \in [-1,1]$ and a corresponding observed output $v_k(t)$, the total capacity is defined as:
	\begin{equation} 
		MC = \sum_{k=1}^{\infty}C(y_k)\end{equation} 	
	Notably, Jaeger demonstrated that for i.i.d. data, the $MC$ is theoretically bounded by the number of observables, $MC \le N$. This upper bound is achievable with a linear reservoir. 
	
	Beyond Linear Memory Capacity ($MC$), Dambre et al. \cite{Dambre2012} introduced the Information Processing Capacity ($IPC$) framework to evaluate a system's ability to reconstruct both linear and non-linear combinations of past inputs. The \(IPC\) considers a (theoretically-infinite) set $\kappa$ of orthogonal targets of the entire input history, constructed from Legendre polynomials which are orthogonal over the input range [-1,1]. 
    
	Each target function $y_k$ is characterized by a specific total degree $d$, defined as the sum of the degrees of its constituent polynomials. This allows the total capacity to be decomposed by degree, where the capacity at degree $d$, denoted as $C_d$, is computed as the sum over all polynomial functions of that exact degree:
	\begin{equation}
		C_d = \sum_{k \in \kappa,\ \text{deg}(y_k) = d} C(y_k).
	\end{equation}
	
	Consequently, the total capacity can thus be expressed as:
	 \begin{equation} 
	 	IPC = \sum_{d=1}^{\infty} C_d.
	\end{equation}

Notably, the first-degree capacity $C_1$ corresponds directly to the linear $MC$, representing the linear special case of this generalized framework. For i.i.d. inputs, the total capacity is theoretically upper-bounded by the number of linearly independent system observables $N$ ($IPC \le N$). This metric is utilized in the subsequent sections to provide a comprehensive assessment of the system's total computational power.
	
The specific training configurations associated with these two metrics are defined as follows. For the $MC$, the number of training samples is $20,000$, the maximum delay is $300$, and the numerical threshold is $10^{-2}$. For the $IPC$, the number of training samples ($n_{\text{train}}$) is $2 \times 10^6$ if $N_{x} = 20$, $3 \times 10^6$ if $20 < N_{x} \leq 50$, and $5 \times 10^6$ if $N_{x} > 50$.

	\section{\label{sec:level3} Results}
	\subsection{Linear and non-linear memory evaluation}

			\begin{figure*}
		\centering
		\includegraphics{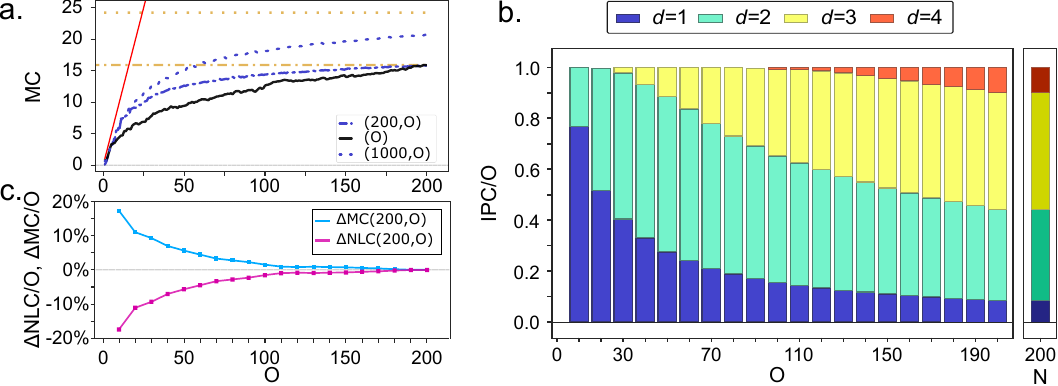}
		\caption{
			\textbf{(a)} $MC$ as a function of the number of observables $O$. The solid black curve represents  $(O)$. The blue dash-dotted and dashed curves correspond to $(200,O)$ and $(1000,O)$, respectively. Horizontal orange dash-dotted and dashed lines indicate performance for the $(200)$ and $(1000)$ sources cases, respectively, while the solid red line marks the theoretical upper bound ($MC = O$).    
    		\textbf{(b)} Evolution of the $IPC$ distribution when projecting from $(200)$ into progressively larger readout spaces, normalized by their respective dimensions (with the baseline $(200)$ distribution displayed on the rightmost edge of the plot).  
			\textbf{(c)} Trade-off between linear memory capacity ($MC$) and non-linear memory capacity ($NLC$) for $(O)$ and $(200,O)$ across different numbers of observables $O$, normalized by $O$ and expressed as a percentage difference.
		}
		\label{fig:wide3}
	\end{figure*}

	We start by computing the MC of the projected and unprojected readout layers. As illustrated by the black curve in Fig.~\ref{fig:wide3}(a), initial simulations of the baseline cases $(N=O)$ show that $MC$ increases as a function of the number of observables ($O \in [1, 200]$). Holding other parameters constant, this scaling behavior is consistent with the findings of Verstraeten \cite{verstraeten2007experimental}.
	
	Furthermore, because the system is non-linear, $MC$ exhibits a sub-linear scaling relationship relative to the increasing number of observables ($MC < O$).

	Subsequently, we analyzed two specific reservoir configurations: $(200)$ and $(1000)$. These configurations yield $MC$ values of 15.91 and 24.22, respectively (represented by the orange horizontal lines on Fig.~\ref{fig:wide3}(a)). We then apply random projection on these states to generate dimensionally compressed readout matrices, $(200,O)$ and $(1000,O)$, made of observables, and then assessed their $MC$, with $O \in [1, 200]$. Notably, the $MC$ for both cases scales non-linearly with the number of observables (blue dotted curves), demonstrating a qualitative scaling behavior analogous to that of the $(O)$ configurations.
	
	Within the specific range of parameters and observables examined in this study, the application of the projection scheme appears to enhance $MC$ relative to the $(O)$ baseline. Empirically, the black curve serves as a lower bound, as the projected configurations maintain equal or superior $MC$ throughout the tested intervals. 
	
	Furthermore, for a fixed number of observables $O_0$, the $\text{MC}$ of $(N, O_0)$ configurations appears to scale positively with $N$. This relationship is evidenced by the superior value of $MC$ of $(1000,O)$ compared to $(200,O)$ (long dots vs short dots blue curves). Despite sharing identical readout dimensions, the systems yield distinct $MC$ values governed by the dimensions of their high dimensional source reservoirs.
	 
	Interestingly, the $MC$ of $(N,O)$ cases never exceeds that of $(N)$, indicating that random projection does not generate additional capacity but rather performs a lossy transformation of its source capacity. Consequently, while the $(N,O)$ $MC$ is inherently bounded by the total information available in the $(N)$ state space, a larger initial $MC$ "budget" (larger $MC$ value in the $(N)$) seems typically to yield higher capacity in the dimensionally compressed case $(N,O)$.
	
	These results suggest that, at least within the scope of these experiments, the projection method more effectively leverages the ability to reconstruct past inputs for a given number of observables than a $(O)$ configuration. Mathematically, these different cases seem to follow a sublinear power law, $MC \propto O^{\alpha}$ (with $0 < \alpha < 1$), reflecting a characteristic saturation of $MC$ as the number of observables increases. Moreover, the retained $MC$ seems to increase monotonically with N, the dimensionality of the source reservoir. Expanding the source size increases the scaling exponent, following the hierarchy $\alpha_{\text{unprojected}} < \alpha_{N=200} < \alpha_{N=1000} < 1$. Consequently, a larger N seems to delay this sublinear saturation, allowing $(N,O)$ to track the ideal linear upper bound over a wider range of observables, especially in the regime of strong compression.

	Nevertheless, the gains in $MC$ facilitated by random projection, relative to the unprojected case, with an equivalent number of observables, entail an inherent trade-off. As established by the $IPC$ framework \cite{Dambre2012}, a reservoir’s total capacity is distributed into linear and non-linear contributions. To investigate how projection reconfigures these distributions, we analyze the partition of these capacities for the $(200)$ case, which operates at full capacity ($IPC=200$), alongside various $(200,O)$, spanning $O \in [10, 200]$ in steps of 10.
	
	To visualize these various profiles, Fig~\ref{fig:wide3}(b) illustrates the $IPC$ across these cases as a function of the normalized observable count. The profiles transition from left to right, with the $(200)$ case on the rightmost side of the plot.
	A primary observation is that with projection, the different cases exhibit various degrees of $IPC$. Given that the projected cases are generated from a $(200)$ possessing multiple degrees of $IPC$ through a linear process of matrix multiplication, this behavior is entirely consistent. Secondly, it can be observed that these various cases consistently achieve ($IPC=O$). This implies that, when starting from a full-capacity system, the projection process generates cases that inherits at least this full capacity from the source reservoir. Moreover, if we consider only the first degree (which corresponds to the $MC$), it constitutes a substantial fraction in highly compressed readout cases before undergoing a gradual decay in total memory, mirroring the sub-linear profile observed in Fig.~\ref{fig:wide3}(a). Consequently, its non-linear counterparts follow a similarly non-linear trajectory, as the cases are bounded by the observables counts.
	
	Accordingly, the distribution of these capacities deviates both from that of ($200$) and across the different $(200,O$) configurations. This phenomenon is especially pronounced in regimes of strong compression (low-dimensional embeddings, or small $O$), where the systems exhibit distinct capacity distributions. Conversely, as the compression ratio decreases and the system shifts toward higher-dimensional embeddings, these distributions tend to converge. Furthermore, as the number of observables increases, higher-degree capacities emerge sequentially, each governed by a specific critical threshold.
	 	
	This is consistent with the fact that increasing the number of observables expands the observables space, thereby facilitating the projection of higher-order polynomials required for degree construction. This behavior is framed by the JL lemma: a higher target dimension imposes fewer constraints on the transformation, minimizing distortion and yielding a projected space closer to that of the source one. These observations are consistent with \cite{cwvm-s53p}, where they showed that expanding a readout layer allows higher degrees of $IPC$ to emerge. Finally, because the dynamics are fully determined by the ($IPC$) of ($N$), the ($N,O$) never exhibit a degree that is not already present in the source space, nor do they achieve a higher capacity.
	
	We now compare ($O$) cases $IPC$ with their ($200,O$) counterparts. To facilitate this analysis, we define the Non-Linear Capacity ($NLC$), which represents the $IPC$ excluding the linear $MC$. To compare between the two cases, we use the following notation for clarity
    \begin{equation}
        \Delta MC(N,O) = MC\big((N,O)\big)-MC\big((O)\big),
    \end{equation}
    \begin{equation}
        \Delta NLC(N,O) = NLC\big((N,O)\big)-NLC\big((O)\big).
    \end{equation}
    Fig.~\ref{fig:wide3}(c) shows the differences in $\Delta MC\big((200,O)\big)$ and $\Delta NLC\big((200,O)\big)$ normalized by \(O\) as a percentage.. 
	
	The first observation of Fig.~\ref{fig:wide3}(c) is that the two curves have the same absolute value for all considered numbers of observables (both cases are at full $IPC$, therefore $\lvert \Delta MC \rvert = \lvert \Delta NLC \rvert$). This indicates a direct trade-off between linearity and non-linearity, suggesting that the projection method acts as a mechanism for tuning the balance between $MC$ and $NLC$. 
	This normalized difference is largest for small numbers of observables, which is consistent with the gap between the blue and black curves in Fig.~\ref{fig:wide3}(a).

	As the number of observables increases, both curves progressively converge toward the horizontal dotted line at zero. In the limiting case where the $(N,O)$ case reaches the same dimensionality as its source reservoir, i.e. $(N,O) \rightarrow (N,N)$, $(N,O)$ and ($N$) exhibit the same $IPC$ distribution, resulting in the vanishing of the differences between $MC$ and $NLC$ of both cases.
	
	In short, the projection process enables small readouts to generate an amount of $MC$ that small reservoirs intrinsically cannot produce, with all other hyperparameters held constant. This mechanism effectively shifts the linear–non-linear trade-off toward linear memory, though at the cost of non-linear components ($NLC$). 	
    
	However, as the embedding size expands, the constraints imposed by the JL lemma become negligible, ensuring that the projected space closely preserves the geometry of its source reservoir. Since the capacity distribution of this large source reservoir already converges toward the baseline configuration, this geometric preservation ultimately causes the IPC capacities per degree to converge across all cases ($\lim_{O \to 200} \Delta MC(200,O) = 0 \quad \text{and} \quad \lim_{O \to 200} \Delta NLC(200,O) = 0$).
	
	Ultimately, these findings demonstrate that the projection framework successfully preserves essential features of the source reservoir, even within larger embedding spaces. This robust preservation under compression provides a crucial foundation for the task-based benchmarks evaluated later in Sec. III.C.

	\subsection{Physical hyperparameter tuning}

	\begin{figure*}
		\centering
		\includegraphics{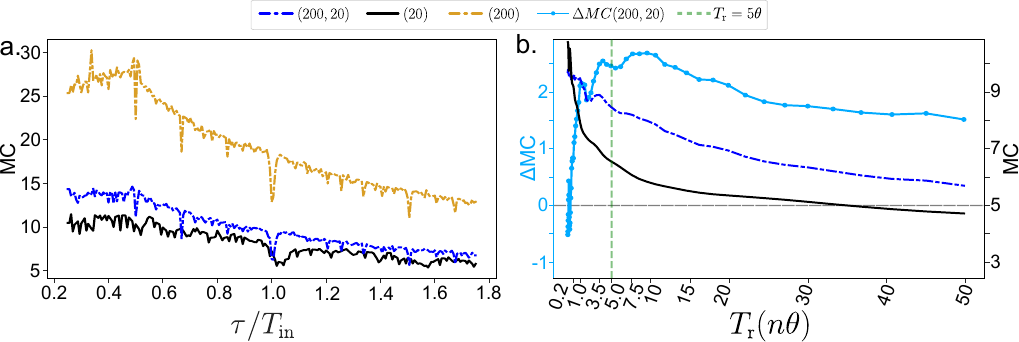}
		\caption{
			\textbf{(a)} $MC$ for different mismatches between $\tau$ and $T_{in}$, shown for three cases: $(200)$, $(20)$ and $(200,20)$
			\textbf{(b)} Memory capacity difference ($\Delta MC$) as a function of the response time $T_{r}$, compared between $(20)$ ($200,20$). The corresponding absolute capacity values are displayed for reference (right axis).		
		}
		\label{fig:wide4}
	\end{figure*}

	 This section examines how model dependencies affect the processing abilities of both ($O$) and ($N,O$) configurations. The analysis first addresses hyperparameters related to the system’s time scales, before providing a comprehensive investigation of the linear and non-linear processing capacities when varying the input scaling factor $ \gamma$.
	\subsubsection{Time \protect\& Synchronization}

	A fundamental design consideration for TDRCs involves the timing of input injection ($T_{in}$) relative to the delay loop ($\tau$). The selection between the so-called \textit{synchronous} ($T_{in} = \tau$) and \textit{asynchronous} ($T_{in} \neq \tau$) operating modes significantly affect the underlying reservoir dynamics. This relationship has been rigorously examined by Stelzer et al.\cite{STELZER2020158}, where they found synchronous operation results in high degrees of informational redundancy within the system, leading to lower values of $MC$ compared to asynchronous cases.

	Fig.~\ref{fig:wide4}(a) tracks the evolution of $MC$ as a function of the $T_{in}/\tau$ ratio for three configurations: $(20)$, $(200)$ and $(200,20)$. As illustrated on the graph in Fig.~\ref{fig:wide4}(a), the $MC$ of the $(200,20)$ case (blue line) closely mirrors the memory trends of the $(200)$ one (orange line), suggesting that the  $(200,20)$ case functions as a down-scaled version of the state it is projected from. Notably, the $(200,20)$ consistently outperforms the $(20)$ case in terms of $MC$. This indicates that the $MC$ for a given number of observables can be substantially enhanced through the application of the projection method, and is not being restricted to specific mismatch ratios. For the asynchronous case $T_{in}=0.9\tau$ and the other specific set of hyper-parameters , the $(200,20)$ configuration yields an improvement of up to 3 units of $MC$ relative to the $(20)$ one.

	Within the explored hyperparameter space, the synchronous regime consistently provides the lowest MC for both  $(200,20)$ and $(20)$ cases. This drop in $MC$ aligns with the findings of \cite{STELZER2020158} mentioned above.
	Conversely, the asynchronous regime augments $MC$, a key requirement for many target applications. The asynchronous regime $T_{in}=0.9\tau$ serves as the default operational mode for the following analyses unless explicitly stated otherwise. 
	
	An additional parameter is the system’s characteristic response time, denoted as $T_{r}$. Within the framework of the Ikeda system, this variable functions as the time constant governing the internal dynamics. The ratio between $T_{r}$ and the virtual node duration $\theta$, determines the connectivity strength between successive virtual neurons. Specifically, the ratio $\theta / T_{r}$ dictates the system's coupling regime: a large ratio results in rapid neuronal responses that diminish inter-neuronal influence, whereas a smaller ratio induces a temporal averaging effect/forward coupling of the response of one node to the successive ones. Fig.\ref{fig:wide4}(b) compares $(20)$, with $(200,20)$, for an increasing ratio.	

	By modulating this ratio, it is observed in Fig.\ref{fig:wide4}(b) that the projection method manages to typically yield a higher or equal $MC$ across the entire range of $T_{r}$. A more granular analysis of this curve reveals an optimal operating point where the $MC$ of the  $(200,20)$ case significantly exceeds that of the  $(20)$ one, providing an additional $2.5$ $MC$ units. Notably, this peak aligns with the selected value of $T_{r} \approx 5\theta$, a value which allows to enhance inter-neuronal coupling while mitigating excessive temporal averaging \cite{Ortin:20}. Within the short response time regime ($T_r \ll \theta$), the performance discrepancy between both configurations disappears. This is evidenced in Fig.\ref{fig:wide4}(b) by the close convergence of the thin dotted blue and black lines (the true $MC$ values), alongside $\Delta MC$ vanishing toward the horizontal axis for large ratios. Interestingly, this limit yields the highest absolute $MC$ values for both configurations. This finding aligns with \cite{Ortin:20}, which demonstrated that a smaller time constant maximizes $MC$ in asynchronous systems, a state commonly referred to as the map limit regime.

	\begin{figure*}
		\centering
		\includegraphics{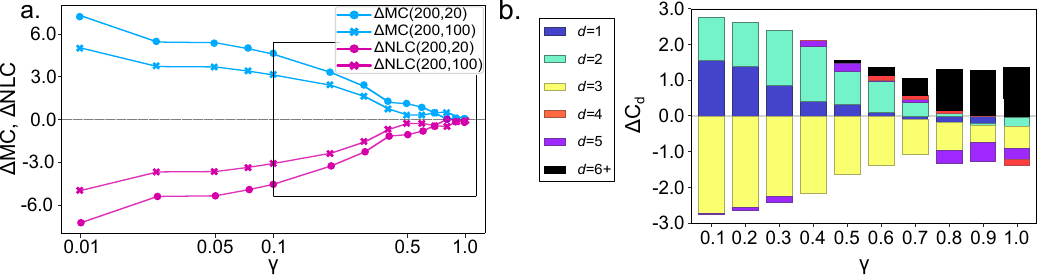}
		\caption{
			\textbf{(a)} $MC$ and $NLC$ differences between $(200,O)$ and $(O)$, as a function of the input scaling factor ($\gamma \in [0.01, 1.0]$). Curves show the discrepancy for $O=20$ and $O=100$.
			\textbf{(b)} Information Processing Capacity ($IPC$) differences per degree as a function of the input scaling factor ($\gamma \in [0.1, 1.0]$) between $(200,O)$ and $(O)$.
		}
		
		\label{fig:wide5}
	\end{figure*}

	The observation that $\Delta \text{MC}$ approaches zero for short response times relative to node separation suggests that the performance gain yielded by the projection is fundamentally contingent upon the existence of shared information, or temporal coupling (inertia), between virtual neurons. Following the framework established by \cite{cwvm-s53p}, $MC$ is intrinsically linked to the underlying correlation structure of the reservoir states. In TDRC, the map-limit regime provides highly uncorrelated neurons. Although $(N)$ may possess $N$ independent directions, any linear projection onto an $M$-dimensional target subspace (with $M \le N$) remains strictly bounded by the rank of that target space. Consequently, since the projection is a linear transformation, the $(N,M)$ states may preserve this lack of correlation without generating new dynamical dependencies or altering the effective dimensionality of the information. This structural limit explains why the difference of $MC$ vanishes, as these two configurations, $(N,M)$ and $(M)$, can be expressed as structurally identical and nearly orthogonal state spaces. Conversely, the performance discrepancy also diminishes in the opposite limit of a large response time relative to the inter-neuronal timescale ($T_r \gg \theta$). In this regime, however, the convergence is not driven by the shared structural limits described above, but rather by a dominant averaging effect of both $(N)$ and $(M)$ cases that ultimately causes the absolute performance of both configurations to decay toward zero.

	In summary, while TDRCs facilitate an enhanced $MC$ through projection for given number of observables, the magnitude of this gain is heavily dependent upon the operational configurations. If performance is already limited within ($N$), there is no compensation offered by the projection to mitigate an improper choice of $T_r$ or $T_{in}$ in dimensionally reduced cases ($M < N$). Instead of recovering missing information, the projection simply propagates a poorly conditioned state space.
	
	\subsubsection{Impact of the input scaling factor}

	The next parameter under evaluation is the input scaling factor, $\gamma$, whose impact is examined across several values, from a weakly input-driven regime ($\gamma=10^{-2}$) to a strongly input-driven regime ($\gamma=1$). Fig.~\ref{fig:wide5}(a) illustrates the difference between $(200,O)$ and $(O)$ regarding $MC$ and $NLC$ for 20 and 100 observables. First thing to notice is that the $(200,O)$ case yields a significant surplus in $MC$ relative to the $(O)$ cases, evidenced by the differential curves (blue curves) that almost consistently remain above the zero-baseline. This performance enhancement is most pronounced in the weakly system-driven regime (low $\gamma$), where the difference between the $(200,O)$ and $(O)$ cases reaches its maximum. However, as the input scaling factor increases, these curves undergo a monotonic decline and eventually converge toward the low dimensional performance profile as $\gamma$ approaches unity.

	This trend appears to be influenced by the total $MC$ available within ($200$). At low $\gamma$, ($200$) cases act as rich sources from which the ($200,O$) cases can efficiently extract a substantial $MC$ surplus. This higher amount of $MC$ in a weakly system-driven regime aligns with \cite{Ortin:20}, where it was demonstrated that the input scaling factor can be adjusted to fine-tune the $MC$ value by keeping the system closer to a linear regime. Because the ($O$) case is inherently limited in its ability to generate such high levels of $MC$, the projection serves as a mechanism to unlock and provide these higher $MC$ values. Conversely, as the systems become more strongly driven, the intrinsic memory of ($N$) cases are depleted. Consequently, at high $\gamma$, ($N$) cases generate such low $MC$ values that the ($N,O$) cases have nearly nothing left to transmit, rendering the $MC$ gain negligible and causing the performance to converge with the ($O$) cases. These results indicate that while the projection scheme is a powerful multiplier for linear memory extraction, its utility upon memory is strictly contingent upon the system operating below the threshold of excessive input strength.

	Regarding nonlinearity, the expected linearity-nonlinearity trade-off is clearly reflected in the $\Delta \text{NLC}$ trajectories, which mirror the behavior of $\Delta \text{MC}$ relative to the zero-baseline. This inverse relationship stems from the conservation of total capacity; across all configurations, both the projected and low-dimensional cases consistently achieve full $\text{IPC}$. Consequently, any gain in $\text{MC}$ is mathematically offset by a corresponding loss in $\text{NLC}$, a balance visually demonstrated by the sum of deviations (represented by the two lines  blue and magenta) equaling zero. Notably,  at $\gamma > 0.5$, this difference ceases to be discernible.

	A more granular examination of the $\text{IPC}$ distribution is provided in Fig.~\ref{fig:wide5}(b), which decomposes the capacity differences by degree for the $(200,20)$ and $(20)$ configurations across $\gamma$ values ranging from $10^{-1}$ to $1$. Since the first-degree $\text{IPC}$ is mathematically equivalent to the $\text{MC}$ (represented here by the blue rectangles), this breakdown naturally echoes the differential trends previously observed in the curves of Fig.~\ref{fig:wide5}(a).
		
	A closer look reveals that at low input gains, the $(200,20)$ configurations enhance not only the first-degree component but also the second-degree $IPC$. In contrast, the ($20$) cases exhibit a higher proportion of third-degree components. These distributions are highly sensitive to the input scaling, suggesting that the projection method efficiently captures the lower-order features of ($200$). These features are inherently more stable, which likely enhance their robustness through the application of the projection. This effect may also be due to the high baseline values of these lower-degree components in ($200$) at low input scalings. As input scaling increases, this trade-off diminishes and higher-degree components emerge more rapidly in the ($200,20$) cases than in the ($20$) cases. This remains consistent with the principle that a larger reservoir, when strongly driven, generates a broader range of non-linear degrees. Indeed, the projection process enables a partial manifestation of these higher-degree $IPC$ components, capturing features that a low-dimensional reservoir cannot generate on its own.
	
	Ultimately, while projection facilitates a significant increase in lower-degree $IPC$ at small and intermediate input scalings, it seemingly does so at the expense of higher-order capacities. 
	
	Because the total information contained within a fixed-size readout layer is fundamentally constrained by the theoretical upper bound of the $IPC$, this trade-off principle can be extended to various other system parameters. Therefore, given that ($N,O$) primarily inherit the lower-degree components of the $(N)$ cases, the goal should be to optimize the $(N)$ IPC profile. Fine-tuning this underlying distribution ensures that random projection maximizes task-relevant information for a given number of observables, as high-degree IPC terms are typically superfluous.

	\subsection{Effect on a real task: NARMA performance}

	\begin{figure*}
		\centering
		\includegraphics{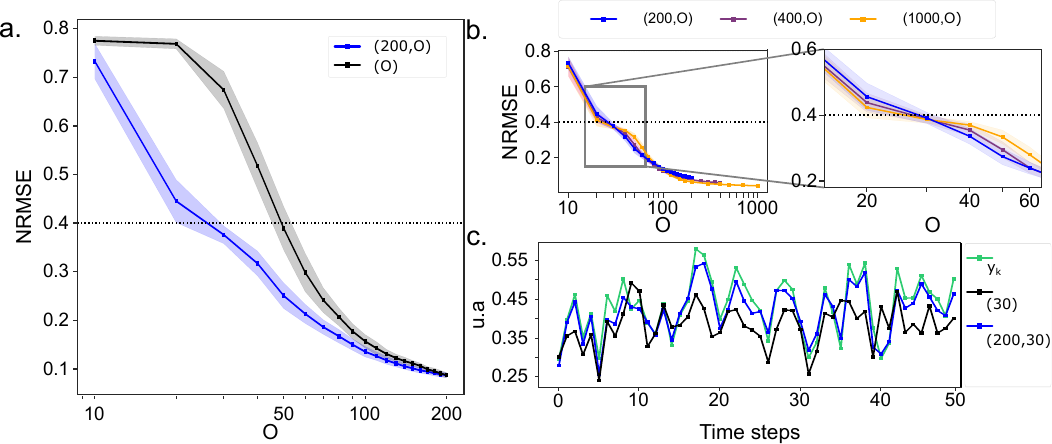}
		\caption{
			\textbf{(a)} NARMA10 performance of NRMSE for increasing number of observables, $(O)$ cases vs $(200,O)$ cases.		
			\textbf{(b)} NARMA10 performance of NRMSE for increasing number of observables, three different cases of source dimensions: $(200,O)$, $(400,O)$ and $(1000,O)$.
			\textbf{(c)} Time trace over 50 steps for 30 observables, comparing $(30)$, $(200,30)$ , and the target.	
            The shaded regions in a,b denote the standard deviation.
		}
		\label{fig:wide6}
	\end{figure*}

	The NARMA10 (Non-linear Autoregressive Moving Average) task represents a standard benchmark for evaluating highly nonlinear sequential processing, specifically designed to challenge a reservoir's ability to model complex temporal dynamics \cite{Narma_founder}. In this task, the input sequence $u_k$ is sampled from a uniform distribution $U(0, 0.5)$. The target signal is governed by:
	
	\begin{equation} y_{k+1}=0.3y_k+0.05y_k\sum_{i=0}^{9}y_{k-i}+1.5u_ku_{k-9}+0.1 .\end{equation}
	
	This benchmark is particularly significant because it requires simultaneously the reconstruction of information from ten preceding time steps and the execution of sophisticated nonlinear transformations on both input and feedback signals. Consequently, NARMA10 serves as a suitable use-case to capitalize on our results from the previous section as it tests the trade-off between the system's memory depth and its nonlinear mapping capabilities.
	
	For this task, the training phase consists of $10,000$ steps, the testing phase is evaluated over $3,000$ steps, and a QR decomposition is employed during the training process to optimize numerical stability. The performances are evaluated using the Normalized Root Mean Square Error (NRMSE), defined as: 
    
    \begin{equation} \text{NRMSE} = \sqrt{\frac{1}{n} \sum_{k=1}^{n} \frac{(y_{k} - \hat{y_{k}})^2}{\tilde\sigma_y^2}}.\end{equation} 
    
    where $y_{\text{k}}$ denotes the target values, $\hat{y_{k}}$ represents the predicted values, and $n$ is the total number of test samples. $\tilde{\sigma}_y^2$ is the sample variance estimated from the input data by $\tilde{\sigma}^2 = \frac{1}{n-1}\sum_{k=1}^n\left(y(k)-\overline y\right)^2$ with $\overline y$ the sample estimate of the mean of the input data.

	Fig. \ref{fig:wide6}(a) presents a comparative performance analysis for the NARMA10 task, evaluating the NRMSE as a function of the number of observables. The benchmark compares $(O)$ configurations against $(200,O)$ ones (averaged over 60 independent trials, with shaded areas representing the standard deviation). The $(200,O)$ configurations (blue line) consistently outperforms the $(O)$ ones (black curve), particularly in the highly-compressed regime. Furthermore, while the $(200,O)$ cases exhibits a rapid error reduction as observables are added, the $(O)$ ones maintain significantly higher error levels, only showing improvement at much higher observable counts. For a sufficiently large number of observables, the performance gap narrows, as both architectures converge toward a similar error floor.

	When interpreted through the lens of $IPC$, the observed performance disparity may be due to the distinct distributions of computational degrees characterized in the preceding sections. Specifically, while both configurations eventually achieve $IPC$ saturation, their capacity allocation across degrees differs fundamentally. This divergence becomes especially pronounced under high compression. These results suggest that the $(200,O)$ configurations provide an IPC profile better suited for the NARMA task than a standard reservoir operating under identical readout constraints.
	
	To further substantiate this interpretation, a direct link can be established between the requirements of the NARMA10 benchmark, the $\text{MC}$, and the second-degree $\text{IPC}$. The NARMA10 task imposes an inherent memory constraint, necessitating an $\text{MC}$ of at least 10 because its formulation explicitly relies on a 10-step input history from $u_k$ down to $u_{k-9}$. Crucially, it also couples these delayed inputs via quadratic cross-products (such as $u_k u_{k-9}$), making the task heavily dependent on second-degree $\text{IPC}$ components. 	
	Our previous results indicate that the projection method tends to reach earlier these critical $MC$ and second-degree $\text{IPC}$ thresholds using a smaller number of observables compared to the $(O)$ cases. This efficiency in accumulating the two first degrees of $\text{IPC}$ may explain the superior performance in the low-observable regime. 
	
	A previous study \cite{linear_nonlineat_trade_off} demonstrated that, for a fixed network size, the ratio of linear-nonlinear neurons can be finely tuned to optimize performance on the NARMA task. By balancing this linearity-nonlinearity trade-off, they also achieved performance enhancements, but by assigning distinct functional roles to specific sub-populations of neurons. In Fig.\ref{fig:wide6}(a), the NRMSE of $(200,O)$ cases exhibit a rapid and significant decline until it reaches the linear reservoir baseline of approximately $0.4$. This sharp improvement suggests that the projection method's ability to capture at least the essential temporal dynamics is much more efficient than ($O$) cases with similar number of observables. 
	
	While a certain amount of $MC$ is required as a baseline to attain acceptable performance, merely increasing this linear component past a specific threshold fails to drive further improvement. Instead, reaching better performances on the NARMA task seems to necessitate stronger nonlinear processing capabilities.

	To carry a deeper analysis on the impact of size of the source reservoir, Fig.~\ref{fig:wide6}(b) compares the performances of three configurations: $(200,O)$, $(400,O)$ and $(1000,O)$ (represented by blue, purple, and orange curves, respectively). At the lowest number of observables (e.g., at 10), no significant performance disparity is detected between the three configurations. However, as the observable count increases, the $(1000,O)$ cases exhibit the steepest improvement, reaching $\mathrm{NMSE}=0.4$ (corresponding to the best reported performance using a linear reservoir \cite{linear_nonlineat_trade_off}) more rapidly than the smaller configurations. While the $(400,O)$ cases follow an intermediate trajectory with a milder slope, the performance curves eventually cross as the number of observables continues to increase. The $(1000,O)$ and $(400,O)$ cases seem to reach a performance plateau, whereas the $(200,O)$ cases continues to improve. These results highlight the fundamental trade-off between $MC$ and $NLC$ within the total $IPC$. 
	
	Initially, a minimum number of observables threshold is required across all configurations to satisfy basic memory demands. Once met, the projections from larger reservoirs rapidly saturate the required $MC$ but subsequently plateau, as a disproportionate amount of their $IPC$ might be allocated to linear memory, leaving insufficient capacity for other degrees. Conversely, the $(200,O)$ cases avoids over-allocating degrees of freedom to $MC$. Although it crosses the $0.4$ threshold at a higher observable count, it preserves the nonlinear degrees of freedom necessary to track complex dynamics, ultimately achieving superior performances (e.g., at 40 observables). Thus, for a fixed number of observables, the $IPC$ distribution of the source reservoir remains the critical determinant of task-specific performance.
	
	As illustrated in Fig. \ref{fig:wide6}(c), which displays a 50-step time trace for $O = 30$ observables, distinct behavioral differences emerge between the configurations. The blue curve, representing a $(200,30)$ case, aligns much more closely with the green target curve than the $(30)$ case, represented by the black curve. Specifically, the $(200,30)$ case accurately tracks the rapid fluctuations and successfully captures the full amplitude of the target's peaks. In contrast, the $(30)$ case exhibits a significantly attenuated amplitude and fails to reproduce the prominent variations, underscoring the clear advantage of the projection method.

	\subsection{Further discussion}

    In the previous sections, we showed that dimensional compression of the readout can preserve and even improve (in specific operating regimes) the computational capabilities of a photonic reservoir computer while using substantially fewer observables than the corresponding reservoir size. We further demonstrated that these gains are accompanied by a redistribution of the IPC and depend strongly on both the source reservoir dimension and the choice of parameters. While the preceding sections establish these observations empirically, they do not by themselves explain why a compressed readout can outperform an equivalently sized standalone reservoir. In what follows, we summarize and discuss possible qualitative interpretations of these results.
    
    Summarizing the results presented in this work, we note that the capacities per degree of \((N,O)\) are essentially fractions of \((N)\)'s capacities per degree. 
    The capacities of \((N,O)\) can be expressed as
    \begin{equation}
        C^{O}_{d} = \zeta_d C_d^N,
    \end{equation}
    where \(C_d^N\) denotes the degree \(d\) capacity of \((N)\), \(C^O_d\) is the degree \(d\) capacity of \((N,O)\), and \(0\leq \zeta_d \leq1\) is a multiplicative factor. 
    Thus, the IPC can be expressed as
    \begin{equation}
        IPC^O=\sum_d C^{O}_d = \sum_d\zeta_d C_d^N\leq O.
    \end{equation}
    Fully-observable (O) and partially-observable $(N>O, O)$ reservoir computers have the same readout size. However, the recorded state matrices of the two cases may be fundamentally different. For example, in the case of $(20)$, the representation is constrained to a twenty-dimensional phase space (upper bound). Whereas in $(200,20)$, the phase space has a maximum dimensionality of 200, from which $O=20$ observables are extracted. This means those extracted features originate from a potentially richer embedding associated with this larger reservoir. As a result, these observables could exhibit a more favorable state diversity, thereby achieving higher $MC$ despite the same readout dimension. This implies that the performance is determined by the number of observables $O$ and the statistical and dynamical structure of $(N)$ from which they are constructed, as illustrated in our numerical simulations. Indeed, the potential gains in $MC$ or $IPC$ (lower degrees $C_d$) are strongly dependent on the hyper-parameter configuration of $(N)$.
    A larger reservoir can retain information in a larger phase space, allowing a reduced set of $O<N$ projected observables to preferentially select some computational features which may not be produced by an equivalently sized standalone reservoir $(O)$. However, if $(N)$ did not generate the features appropriately, or if \(O/N\approx1\), then the gains from random projection are limited.

    With this view, it is clear that projection is primarily a method of reallocating and redistributing existing computational resources in \((N)\).
    This is confirmed in Fig. \ref{fig:wide3}(b). 
    We note that for higher \(d\), \(\zeta_d\) will diminish faster with respect to increasing compression ratio, while for lower \(d\), \(\zeta_d\) will diminish more slowly.
    As such, for high compression ratios, lower degree capacities `survive' the projection more than higher degrees.

We can also try to explain the increase in $MC$ between the projected cases and the baseline cases using geometric arguments based on the notion of 'Blessing of Dimensionality' \cite{Kainen1997}. 

In the context of high-dimensional geometry, the concentration of measure provides a key structural advantage: randomly chosen vectors become nearly orthogonal with high probability as the dimensionality of that vector space increases. Given two unit vectors $\mathbf{u}$ and $\textbf{v}$ in $\mathbb{R}^N$, and considering the scalar product $f(\textbf{v}) = \textbf{u} \cdot \textbf{v}$ as a $1$-Lipschitz function, this phenomenon is formally described by the following inequality:$$P(|\textbf{u} \cdot \textbf{v}| > \epsilon) \leq 2e^{-\frac{N\epsilon^2}{2}}.$$As noted in \cite{cai2013distributions}, this concentration result provides a precise mathematical characterization of the theorem that all high-dimensional random vectors are almost always nearly orthogonal to each other. In our work, this geometric property ensures that as the reservoir dimension $N$ increases, state vectors are statistically driven toward mutual quasi-orthogonality, which directly facilitates linear separability at the readout layer.
However, the arguments presented here do not fully explain the resulting \(IPC\) distribution from random projection, specifically the preference to lower degree capacities.

	\section{Conclusion}
	
	In conclusion, this paper has demonstrated that for a given output layer size, the processing capabilities of a photonic reservoir computing system are dependent on the dimension of its  high dimensional state space. When a system is constrained by a fixed-size output layer, its performance can be significantly leveraged by operating in a larger  high dimensional state space and subsequently applying a dimension reduction technique. We have shown that, while random projection does not creates additional computational resources, it provides an alternative lower-dimensional view of the phase space of the higher-dimensional source reservoir. Hence, the quality of the projected representation ultimately depends on the information available in the source reservoir.
	
	Nevertheless, this approach is subject to fundamental limitations. Because the total processing capacity remains bounded by the size of the output layer, a trade-off emerges: gaining linear processing capacity inevitably results in a loss of non-linear processing capacity.
	
	Interestingly, however, it seems that random projection can preserve task-relevant features generated within the higher-dimensional phase space. When tested on NARMA-10, for the same number of observables between projected and unprojected cases, the projected case performs much better, specifically within a particular range of compression ratios (a sweet spot). The incorporation of a projection layer and the resulting trade-off open up opportunities for highly flexible implementations of physical output layers in all-optical computing systems.
	
	\begin{acknowledgments}
		
		The authors would like to gratefully acknowledge the financial support of the Conseil of Region Grand-Est through the project PULSE, the Air Force Office of Scientific Research (AFOSR) and Office of Naval Research (ONR) through grant FA8655-22-1-7031, and by the European Union (EU) Chips JU project No. 101194363 (NEHIL).
		
		This work was performed in part using computational resources from the “Ruche Mésocentre” computing center of Université Paris-Saclay, CentraleSupélec and École Normale Supérieure Paris-Saclay supported by CNRS and Région Île-de-France.

	\end{acknowledgments}

	\bibliographystyle{ieeetr} 
	\bibliography{references}

\end{document}